\documentclass{article}
\usepackage{amsmath}
\usepackage{algorithm}
\usepackage{algorithmic}
\usepackage[margin=1in]{geometry}
\usepackage[procnames]{listings}
\usepackage{authblk}

\usepackage{apacite}
\usepackage{graphicx}
\usepackage{subcaption}
\usepackage{color}

\newcommand{\ul}[1]{\underline{#1}}
\newcommand{\mb}[1]{\mathbf{#1}}

\definecolor{keywords}{RGB}{255,0,90}
 
\title{One Model for the Learning of Language}

\author[1]{Yuan Yang}

\affil[1]{School of Computer Science, Carnegie Mellon University, Pittsburgh, 15213, USA}

\begin{document}

\maketitle

\begin{abstract}
A major target of linguistics and cognitive science has been to understand what class of learning systems can acquire the key structures of natural language. Until recently, the computational requirements of language have been used to argue that learning is impossible without a highly constrained hypothesis space. Here, we describe a learning system that is maximally unconstrained, operating over the space of all computations, and is able to acquire several of the key structures present natural language from positive evidence alone. The model successfully acquires regular (e.g. $(ab)^n$), context-free (e.g. $a^n b^n$, $x x^R$), and context-sensitive (e.g. $a^nb^nc^n$, $a^nb^mc^nd^m$, $xx$) formal languages. Our approach develops the concept of factorized programs in Bayesian program induction in order to help manage the complexity of representation. We show in learning, the model predicts several phenomena empirically observed in human grammar acquisition experiments. \color{red}{This is a draft write-up of an undergraduate project. A full journal version is still under preparation.}\footnote{Undergraduate project done with Steven Piantadosi, Department of Brain and Cognitive Sciences, University of Rochester, Rochester, 14604, USA}
\end{abstract}

\section{Introduction}

One of the central debates in language acquisition is whether the syntactic structures of natural language are genetically specified or learned through experience. A key tool in this debate has been the use of formal mathematical analysis in order to determine what learners could or could not logically conclude from the type of data they observe. Perhaps the most famous is \emph{Gold's Theorem} \cite{gold1967language}, which holds that observers of positive examples of even a \emph{regular language} \cite{hopcroft1979introduction} could not identify the set of allowable (``grammatical'') strings with certainty. In linguistics and cognitive science, Gold's proof  was taken to mean that innate constraints must be present for learning to succeed. The theorem gave rise to rich formal theories of what could be learned under similar assumptions and correspondingly what must be innate \cite<e.g.>{wexler1983formal}, although its relevance for cognitive science is not universally accepted \cite{johnson2004gold}. Indeed, more recent work has undermined the view that learners must come with substantial built-in knowledge about the specific structures of language. \citeA{chater2007ideal} describe a theoretical learning framework without tight constraints that could identify the right language still from only positive data. \citeA{chater2007ideal}'s make two key differences from Gold-style analyes: first, \citeA{chater2007ideal} assume that sentences are \emph{sampled} from the parent's generative model, not a worst-case analysis. Second, the model considers \emph{all possible computations} as hypotheses, showing that the space need not be constrained innately for an idealized learner. The core underlying idea of this work is that learners attempt to find simple descriptions of the data they observe, much like work in Minimum Description Length \cite{grunwald2007minimum} and artificial intelligence more generally \cite{hutter2005universal}. This perspective has been further developed to correctly predict the difficulty of constructions \cite{hsu2010logical,hsu2011probabilistic}. 

The present paper applies this perspective to learning the structures supporting natural language that was first presented in Chomsky's ``Three models for the description of language'' \cite{chomsky1956three, chomsky1957syntactic, chomsky1959certain}. Chomsky noted that many dependencies in natural language could be captured abstractly with fundamentally different kinds of computational devices. Some devices require a finite amount of memory (e.g. finite-state machines), some require a stack (e.g. phrase structure rules), and some require even more powerful computational systems (e.g. context-sensitive grammars). These relationships hold true for even very simple versions of natural language structures, including sets of strings---called \emph{formal languages}---over simple alphabets. For example, the simple recursion/iteration allowed by English adjectives (``The beautiful well-dressed young lady'') could be captured by the formal language $\lbrace a^n : n=1,2,3,\ldots \rbrace$ indicating that any number of adjectives ($a$s) could be put together into a valid (sub)string. The language $(ab)^n$ could be seen in, for instance, embedding in lists of determiner-noun pairs (``Bring me two cars, three accordions, and six pickles.'') or in subject-verb pairs of sentences in disourse (``John laughed. Sally cried. Mindy marveled.''). The formal language $\lbrace a^n b^n : n =1,2,\ldots\rbrace = \lbrace ab, aabb, aaabb, \ldots \rbrace$ might characterize the key dependencies in english \emph{if-then} relationships where every ``if'' (an ``a'') must be followed by a ``then'' (a ``b''), as in ``\textbf{If} \textbf{If} Mary cried \textbf{then} John was sad \textbf{then} John cares about Mary.''

In this study, we consider a learning model that, like the one of \citeA{chater2007ideal}, builds in a space of \emph{all computations}. In contrast to Chomsky's \emph{Universal Grammar} (UG), specifying such space, one only need to build in a few operations, a minimal Turing complete set. Results from computer science, complexity theory, and logic have shown that it takes very little formal mechanism to allow Turing-completeness. By building in only such a minimal mechanism---here, lambda calculus \cite{church1936unsolvable, hindley1986introduction}---we are able to construct a theory whose innate assumptions are explicit and extremely minimal. Out of a large, implicitly defined hypothesis space, learners are able to construct \emph{new computations} which correspond to different types of formal language systems. Our model shows that given a single such system, learners could construct or genuinely discover cognitive representations corresponding to finite, regular, context-free, even context-sensitive languages and beyond.

Previous study conducted by \citeA{perfors2011learnability} follows a similar paradigm, where they present a learner who does Bayesian inference over different kinds of grammatical systems, including finite, regular, and context-free grammars. They show that with very little positive evidence in the form of part-of-speech sequences from CHILDES \cite{macwhinney2000childes} learners could infer that natural language requires a context free grammar. However, the key difference between the present work and \cite{perfors2011learnability} is that we do not build in just a handful of possible alternatives; instead, we define an implicit space of hypotheses that is more concise or parsimonious to build in a larger, unconstrained space of hypotheses, allowing the description length of what must be assumed by the model to be very short. As the former has often been criticized to have ``built in'' \emph{more} than alternatives such as UG.

In the next section we describe a computational model that can acquire these kinds of fundamentally different computational structures. We then show that it can learn a variety of formal languages from positive evidence only, and that in doing so it is unproblematic to acquire languages with infinite cardinality, even given finite data. We then show that the model can combine several different computational sub-structures and acquire a grammar of simplified English, including non-context-free language structures described in prior literature. In the last section, we show that the model is able to captures several experimental findings from artificial language learning.

\section{A computational model}

The starting assumption for our model is~\citeA{fodor}'s \emph{Language of Thought} (LOT), the hypothesis that the structures of mental operations are structured and compositional, or language-like,. Following now a growing body of work in Bayesian LOT models \cite{goodman2008rational, piantadosi2012bootstrapping, probmods, piantadosi2011learning, piantadosi2016logical, piantadosi2016four, yildirim2013transfer}, we define a set of operations that compose and have learners consider all compositions of these operations as possible hypotheses. These meanings can be interpreted analogously to short computer programs which ``compute'' strings of part of speech sequences, although the input and output could be different in other domains. Importantly, the specification of these programs is \emph{domain general} meaning that essentially the same setup can be used to explain acquisition in other domains like counting, kinship, cross-modal transfer, and concept learning. In all cases, the task of the learner is to determine which compositions of primitives are likely to be correct, given the observed data. 

\subsection{Hypothesis space}

\begin{table}[t!]
\centering
\caption{Primitive functions used in LOT.}
\label{prim}
\begin{tabular}{lllll}
\hline
\multicolumn{2}{l}{\bf Functions on strings} \\
 $pair(X, Y)$  &   concatenate lists $X$ and $Y$ &   \\
 $first(X)$  &  return the first element of list $X$  &    \\
 $rest(X)$  &  return the list $X$ without the first element& \\
   &  & \\
\multicolumn{2}{l}{\bf Logical functions} \\
 $flip(p)$  &  randomly return true (probability $p$) or false (probability $1-p$) & \\
 $empty(X)$  &  return true if string $X$ is empty, otherwise false & \\
 $if(B,X,Y)$ & return $X$ if $B$ else return $Y$ \\
   &  & \\
\multicolumn{2}{l}{\bf Function calls} \\
 $F_i(s)$  &  call the expression $F_i$ with argument $s$& \\ \hline
\end{tabular}
\end{table}

The primitives that we assume are compoased are motivated by minimalist, functional programming languages like \emph{scheme} \cite{sicp}, which also forms the basis of other LOT work \cite<e.g.>{probmods}. There are 3 classes of primitive functions, shown in Table.\ref{prim}. The first 3 are list operators: \emph{pair}, \emph{first} and \emph{rest}, which respectively join two elements together (somewhat like \citeA{chomsky1995minimalist}'s \emph{merge}), take the first element of a pair, or take the second element. The second class of operators are \emph{flip} and \emph{empty}, which return boolean values. \emph{flip} is notable in that it provides a stochastic element to the grammar, allowing nondeterminism in the generation of structures. \emph{if} allows for conditional expressions. Finally we allowa function to call another function $g_i$---potentially itself---via $function(list)$. We note the similarity between this grammar and others used in the conceptual modeling with a LOT in entirely different domains \cite{piantadosi2012bootstrapping}. While these operations look very simple, they actually permit a surprising range of computations to be expressed. 


\begin{table}[tb!]
\centering
\caption{The grammar used to specify valid compositions of primitives.}
\label{rule}
\begin{tabular}{lll}
\hline
$START \to \lambda x. list$              & $atom \to a,b,c,\ldots$  & \\
$list \to pairs(list, list)$  & $list \to function(list)$ & \\
$list \to rest(list)$         & $list \to if(bool, list, list)$   & \\
$list \to first(list)$        & $bool \to flip(probability)$ & \\
$list \to \emptyset$          & $probability  \to 0.5, 0.6, \ldots 0.9$ & \\
$list \to x$                  & $bool \to empty(list)$ &\\
$list \to atom$ & & \\
\hline
\end{tabular}
\end{table}

Compositions of these primitives and grammar define a set of functions which act as hypotheses for learners. In our analogy to the Library of Babel, these primitives are the characters in the books, and a book (a composition of primitives) is a single hypothesis. The goal of the learner is to walk in the library (hypothesis space) and find the best book (hypothesis) that fits the observed data. For instance, a possible expression built out of these primitives is,
\begin{center}
\begin{tabular}{c}
\begin{lstlisting}[mathescape]
F1(x) = pair( a, if(flip(0.7), $\emptyset$, F1($\emptyset$)))
\end{lstlisting}
\end{tabular}
\end{center}
Here we have named this function ``F1'' and have allowed it to recursively call itself in its own definition via \emph{call}. When this expression evaluates, it concatenates (\emph{pair}s) an ``a'' with either an emptry string ($\emptyset$) or the outcome of calling itself with an empty string as an argument, \lstinline[mathescape]{call(F1, $\emptyset$)}. In terms of a set of strings, this function represents the formal language $\lbrace a, aa, aaa, aaaa, \ldots \rbrace = \lbrace a^n : n = 1,2,3,\ldots \rbrace$. However, this function also gives a \emph{probability distribution} over strings since some (e.g. $a$) are much more likely to be generated than others (e.g. $aaaa$) according to the probability of each \emph{flip}. In particular, this probability distribution this assigns is geometric, $P(a^n) = 0.7^{n-1}\cdot 0.3$. Note, however, that these operations permit a vast array of different types of languages and computations to be specified. 

\subsection{Factorized hypothesis}

One important feature of nearly all working computational systems is that they have a mechanisms to handle complexity, in large part through re-use of subcomputations (For work on learning principled subcomputations to re-use, see \cite{o2015productivity}). One simple way to manage the complexity of hypotheses and efficiently structure search is to allow learners to incrementally build on hypotheses, extending partially working theories in order to cover new data. Here, we introduce a \emph{factorized} hypothesis consisting of a sequence of functions, each of which may be defined in terms of prior functions. This is best illustrated with an example:
\begin{center}
\begin{tabular}{lll}
F1(x) & = & 
\begin{lstlisting}[mathescape]
pair(a, if(flip(0.7), $\emptyset$, F1($\emptyset$)))}} 
\end{lstlisting} \\
F2(x) & = & 
\begin{lstlisting}[mathescape]
if(empty(x), $\emptyset$, pair(first(x), pair(F2(rest(x)), b)))} 
\end{lstlisting} \\
F3(x) & = & 
\begin{lstlisting}[mathescape]
F2(F1($\emptyset$)) 
\end{lstlisting} \\
\end{tabular}
\end{center}
The first fucntion, F1, is simply the $a^n$ language shown above. The second function, F2, is more interesting: it is a function that takes an argument $x$, and recursively concatenates the first element of $x$ with a recursive call to itself, followed by a 'b'.  Because this is recursive on a shortening list (\emph{rest(x)}), one 'b' will be added for each element of $x$. For instance, if we called F2 on the string 'xyz', we would be left with the string 'xyzbbb'. Finally, F3 puts F1 and F2 together, calling F2 with F1 as an argument. Thus, the strings $a^n$ generated by F1 will be passed to F2, which will attach a single 'b' to for each 'a', giving rise to strings of the form $a^n b^n$.\footnote{Note this is just an example, as there are other, simpler ways to generate $a^nb^n$.} In this way, a complex computation can be broken down into pieces that are more manable, each of which might be more easily learned on their own. In the domain of language or grammar learning, such a division of labor is especially elegant since many valid strings have valid substrings, so a model that captures part of the data (the substrings) in F1 can be extended to a model that captures the longer strings as well in F2. Note that in permitting passing arguments to functions, this formamlism permits important types of variable abstraction.

In the implementation, we allow up to 10 different functions to be called in this way. Note that each function can only call itself and prior functions (e.g. F3 cannot call F4, but F4 can call F3 and F4). Also, it is assumed by convention that the output of the function is the final hypothesis called with the empty string as input. 

\subsection{The probabilistic model}

Next, we turn to the specification of a probabilistic model over this space that supports tractable learning and inference. Following prior work using the Bayesian LOT, we convert the grammar of operations to a Probabilistic Context-Free Grammar (PCFG) \cite{manning1999foundations} that specifies the prior probability of any hypothesis. The use of a PCFG automatically penalizes complex hypotheses, fitting the idea that \emph{simplicity} is a core inductive pressure \cite{chater2003simplicity,feldman2000minimization}. 

The goal of a learner is to observe some strings $D$ and infer a posterior on hypotheses, $P(H|D)$. Following Bayes rule, $P(H|D) \propto P(H)P(D|H)$, where $P(H)$ is given by the PCFG times a $2^{-K}$ penalty on having $K$ factorized components (F1, F2, $\ldots$, FK), and $P(D|H)$ is a multinomial likelihood specifying how likely the observed strings are to be generated by $H$. This likelihood include an outlier likelihood penalty of log value $-1000$ for strings in the data that are assigned zero probability by $H$. To manage the uncomputability inherent in this space, hypotheses that do not ``halt'' after a set number of recursive calls---e.g. those that are likely to be in an infinite loop---are assigned a zero prior. 

In inference, most of the posterior probability mass will be held by just a few concise and well-fitting hypotheses. In our implementation (the code can be found on GitHub \cite{piantadosi2014lotlib}), we run 12 Monte-Carlo chains for 50000 steps of the tree-regeneration MCMC method presented in \citeA{goodman2008rational} for varying amounts of data. his algorithm essentially takes a random walk around hypotheses $H$ that is biased towards high probability regions. Over time, this will converge on hypotheses with high $P(H|D)$ and indeed can be shown to correctly be sampling from this posterior distribution. We collect 100 of the highest-posterior expressions found in each chain. These samples are taken as a fixed, finite hypothesis space for the purposes of computing summary statistics and plots. Code is available online in LOTlib \cite{piantadosi2014lotlib}. In each model run, we add to the grammar the required terminal symbols specific to each data set. 


\section{Learning simple formal languages}

We first study the model's ability to learn several simple formal languages that are motivated by the structures present in natural language. Since Chomsky (1956), numerous authors have attempted to pinpoint what type of computational system supports natural language, with much of the debate focusing on what constructions evidence computational requirements more than context free languages. This has motivated attempts to map natural language structures onto very simple formal languages whose computational properties can be fully characterized.

Specifically, we examine the languages $a^n$, $a^nb^n$, $a^nb^{2n}$, $(ab)^n$, $a^nb^nc^n$, and the Dyck language (valid sequences of parentheses). Here, $a^n$ and $(ab)^n$ are regular languages and can be accepted by finite state automation;  $a^nb^n$, $a^nb^{2n}$ and the $Dyck$ language are context free languages, and are recognized by pushdown automation; $a^nb^nc^n$ is a context sensitive language that requires more than a context free grammar. Critically, many of these languages have analogs in natural language synta; and the Dyck language specifies the valid ways in which a sentence could be parsed. 


\subsection{Results}

\begin{figure}
	\centering
    
	\begin{subfigure}[]{0.3\textwidth}
    \centering
	\includegraphics[width=\textwidth]{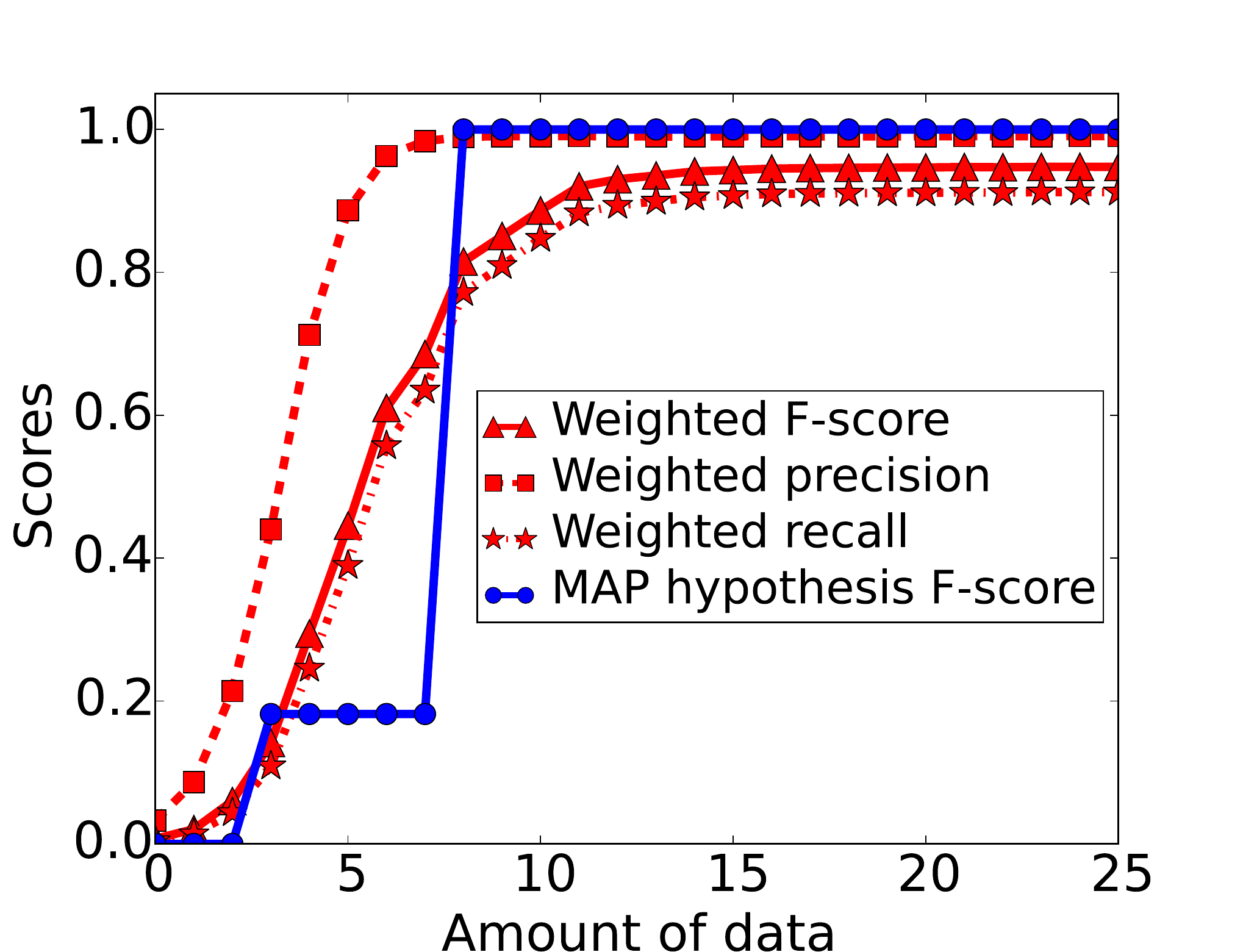}
    \caption{}
    \label{g_an_wf}
	\end{subfigure} \hfill
    \begin{subfigure}[]{0.3\textwidth}
    \centering
	\includegraphics[width=\textwidth]{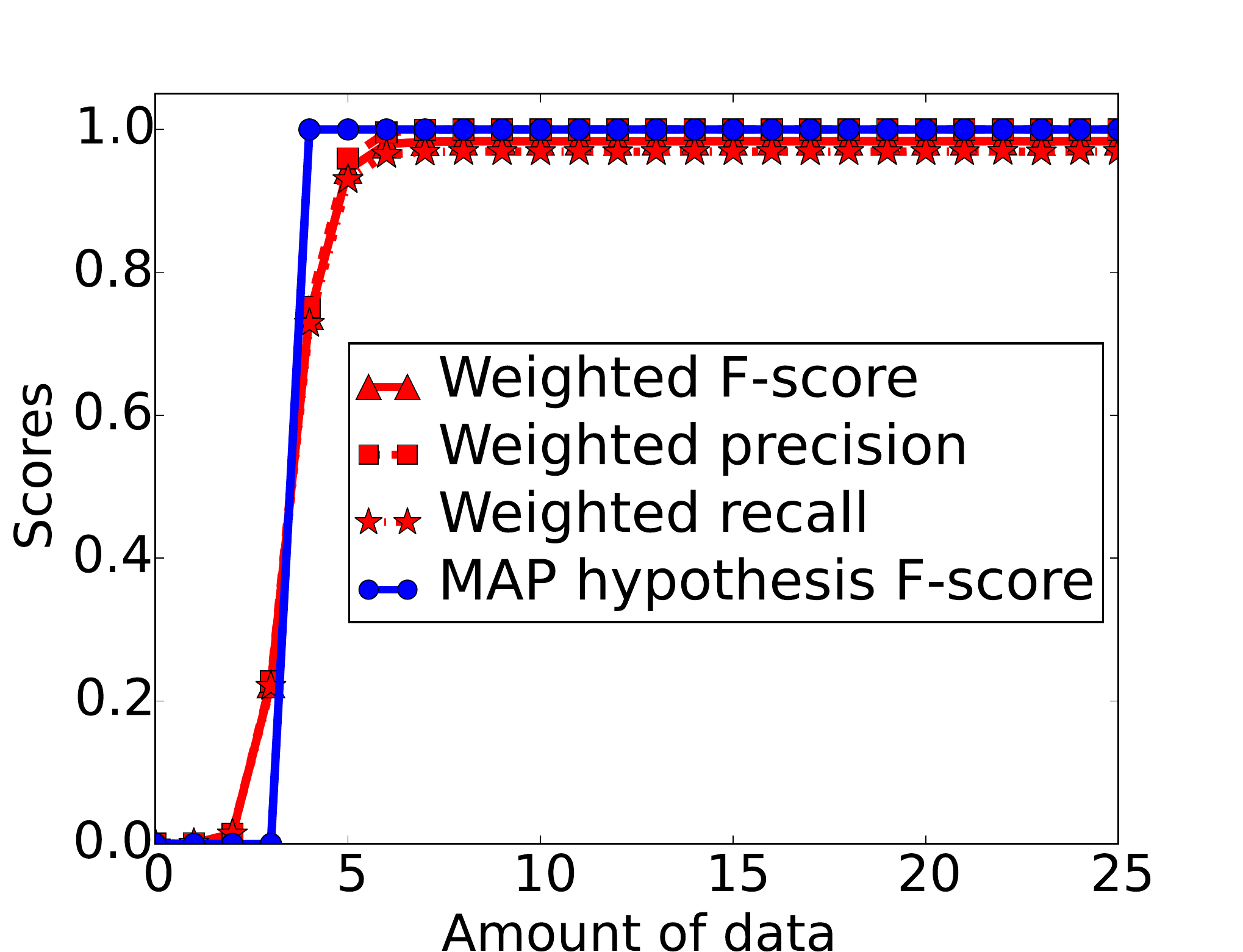}
    \caption{}
    \label{g_abn_wf}
	\end{subfigure}\hfill
    \begin{subfigure}[]{0.3\textwidth}
    \centering
	\includegraphics[width=\textwidth]{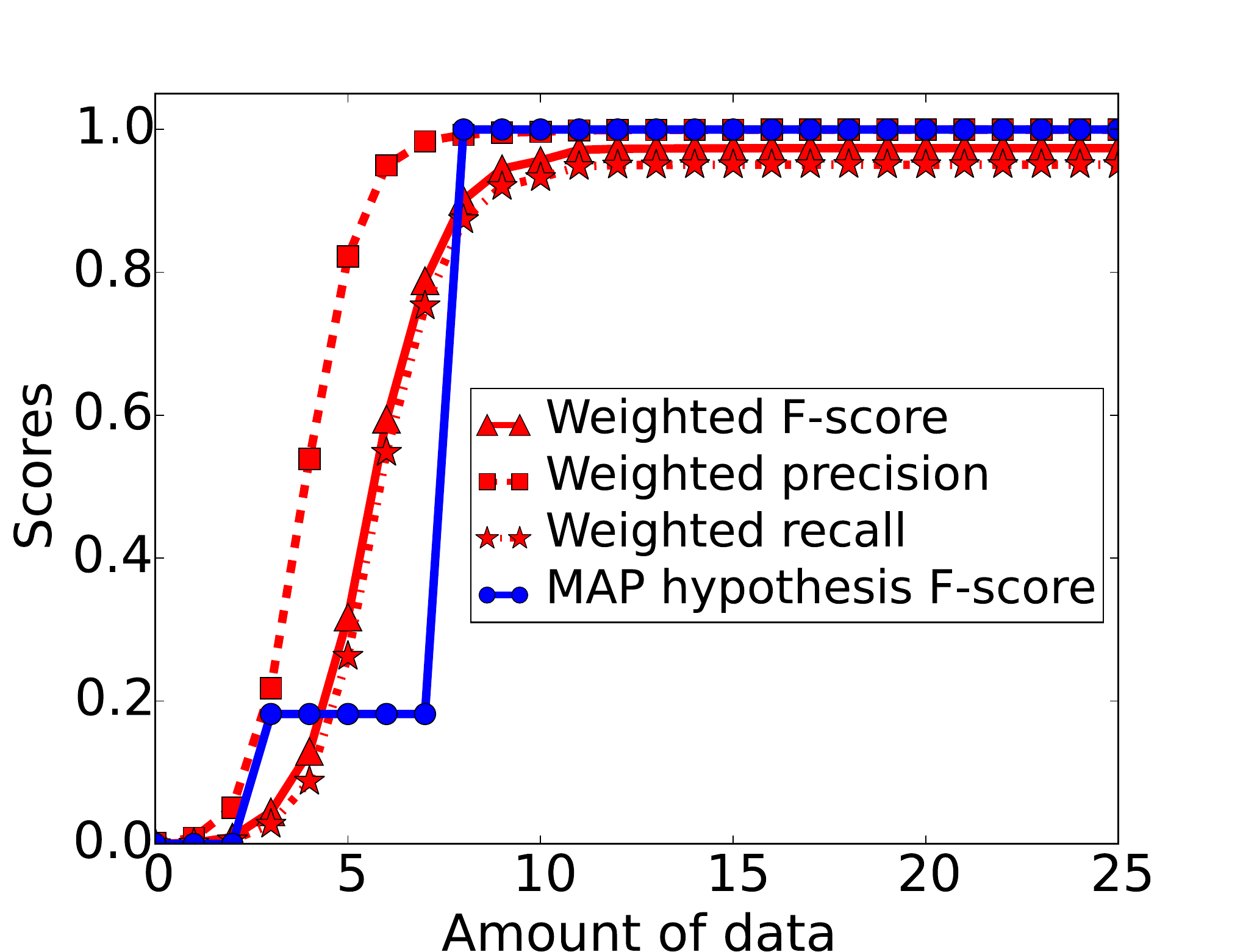}
    \caption{}
    \label{g_anbn_wf}
	\end{subfigure}
    
    \begin{subfigure}[]{0.3\textwidth}
    \centering
	\includegraphics[width=\textwidth]{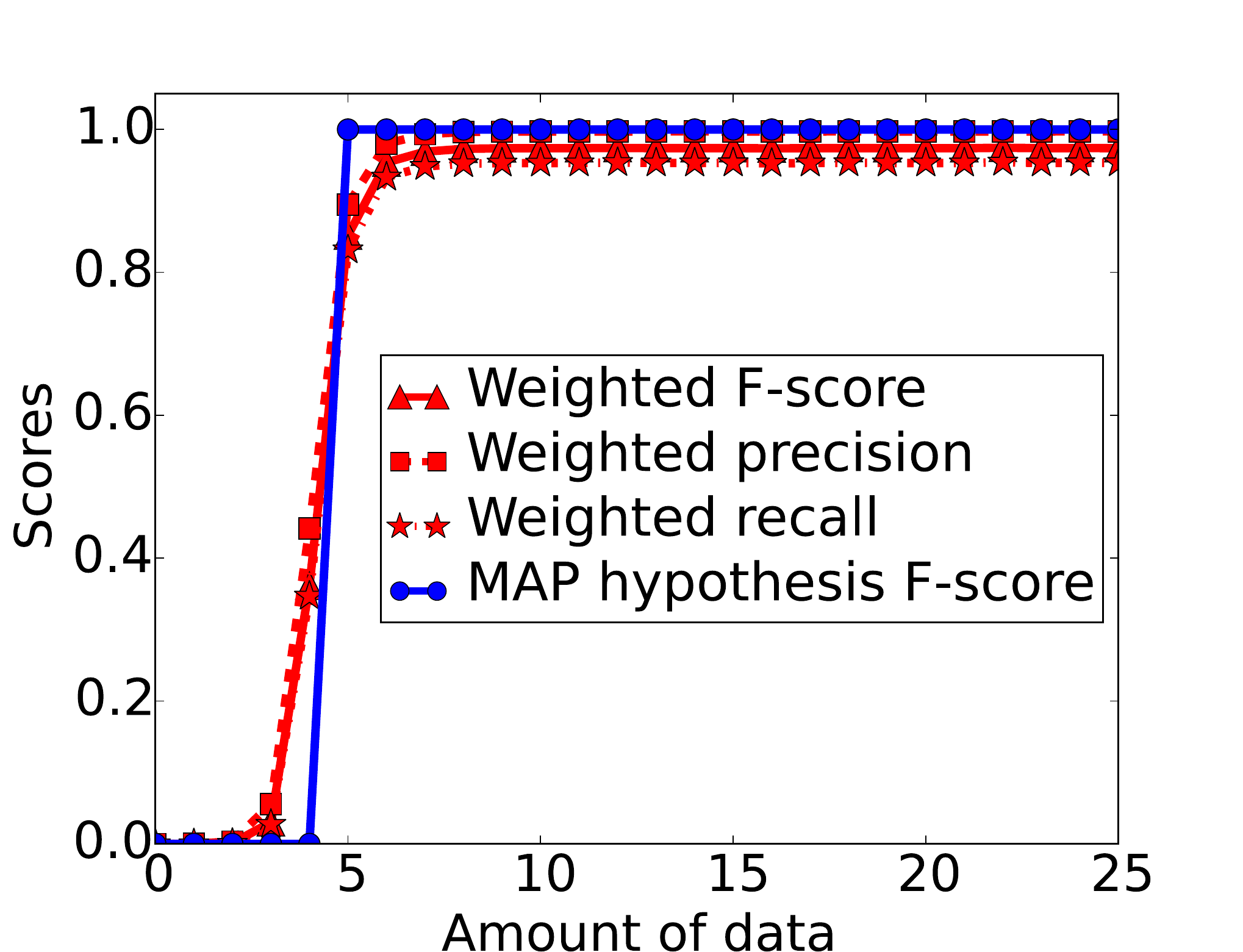}
    \caption{}
    \label{g_anb2n_wf}
	\end{subfigure} \hfill
    \begin{subfigure}[]{0.3\textwidth}
    \centering
	\includegraphics[width=\textwidth]{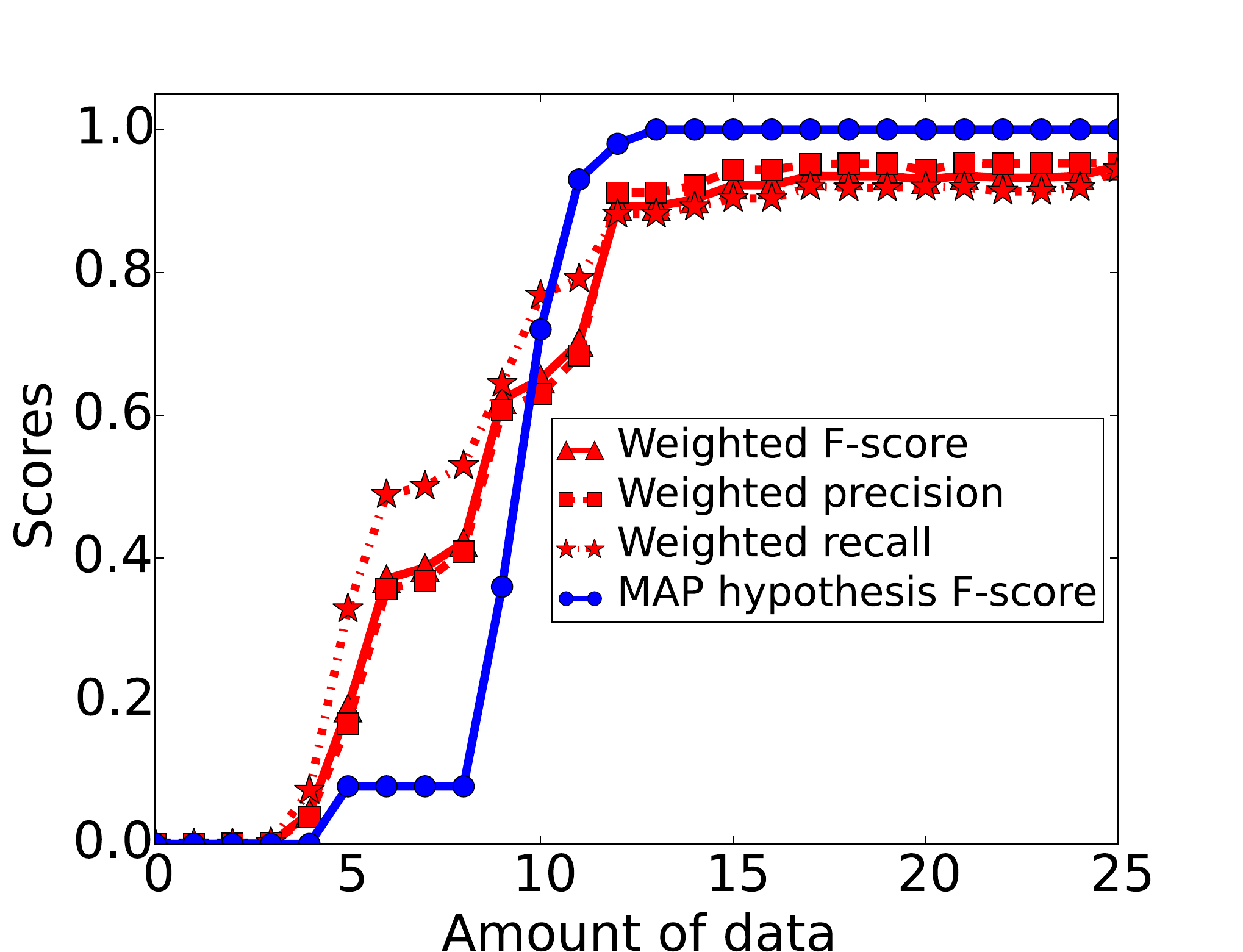}
    \caption{}
    \label{g_dyck_wf}
	\end{subfigure}\hfill
    \begin{subfigure}[]{0.3\textwidth}
    \centering
	\includegraphics[width=\textwidth]{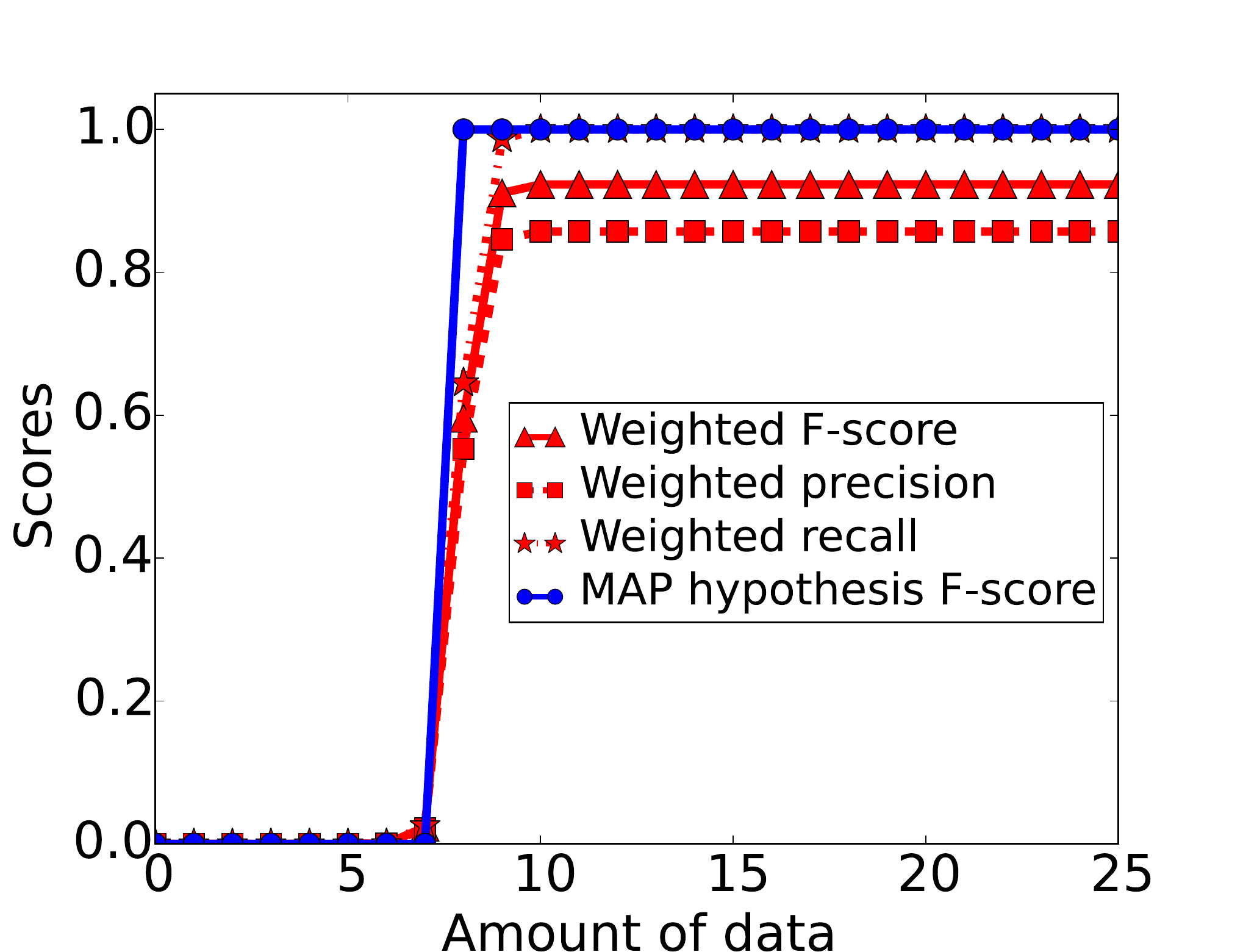}
    \caption{}
    \label{g_anbncn_wf}
	\end{subfigure}
    
    \caption{The graph shows the weighted F-score against the amount of observed data: (a) $a^n$; (b) $(ab)^n$; (c) $a^nb^n$; (d) $a^nb^{2n}$; (e) $Dyck$; (f) $a^nb^nc^n$. The F-score curve for Maximum a Posteriori (MAP) hypothesis is also shown. }
    \label{g_wf}
\end{figure}

Fig.\ref{g_wf} shows the posterior-weighted precision, recall and F-scores of the hypotheses given different sizes of data from these formal languages. Here, the data consist of observations of positive examples of strings (e.g. from $a^nb^n$ we might observe $\lbrace ab, ab, aabb, ab, aaabbb \rbrace$) and is provided to \emph{the same} learning model. The model is able to rapidly acquire the correct kind of computational system for each, often taking only 5-10 samples of strings before the model accurately learns the right system. The implication of this result is that ideal learner with domain-general constraints is able to learn the unobserved grammatical structures and computations that underlie a target set of strings rapidly, and from very little data. To illustrate, \ref{Tab:example_hyp} shows example $perfect$ hypotheses for each type of language.

\begin{table}[htb]
\centering
\caption{Example $perfect$ hypotheses for 6 classes for formal languages.}
\label{Tab:example_hyp}
\begin{tabular}{lllll}
\hline
$a^n$ & $a^nb^n$\\
\begin{lstlisting}[mathescape]
def $F_1$():
    return $pair$($\mb{a}$,$if$($flip()$,$recurse$(),$\emptyset$))
\end{lstlisting}&
\begin{lstlisting}[mathescape]
def $F_1$():
    def $F_x$():
        return $if$($flip()$,$\emptyset$,$F_1$())
    return $pair$($\mb{a}$,$pair$($F_x$(),$\mb{b}$))
\end{lstlisting} \\

\hline
$(ab)^n$ & $a^nb^{2n}$\\
\begin{lstlisting}[mathescape]
def $F_1$():
    def $F_x$():
        return $if$($flip()$,$\emptyset$,$F_1$())
    return $pair$($pair$($\mb{a}$,$\mb{b}$),$F_x$())
\end{lstlisting}&
\begin{lstlisting}[mathescape]
def $F_1$():
    def $F_x$():
        return $if$($flip()$,$\emptyset$,$F_1$())
    return $pair$($\mb{a}$,$pair$($F_x$(),$pair$($\mb{b}$,$\mb{b}$))
\end{lstlisting}\\

\hline
$Dyck$ & $a^nb^nc^n$\\
\begin{lstlisting}[mathescape]
def $F_1$():
    def $F_x$():
        return $if$($flip()$,$\emptyset$,$F_1$())
    return $pair$($\mb{a}$,$pair$($F_x$(),$pair$($\mb{b}$,$F_x$())))
\end{lstlisting}&
\begin{lstlisting}[mathescape]
def $F$():
    def $F_1$():
        return $pair$($\mb{b}$,$\mb{c}$)
    def $F_2$($g_1$):
        def $F_x$():
            return $pair$($pair$($\mb{b}$,$g_1$()),$\mb{c}$)
        return $pair$($\mb{a}$,$if$($flip$(),$g_1$(),$recurse$($F_x$)))
    return $F_2$($F_1$).
\end{lstlisting}\\
\hline
\end{tabular}
\end{table}

\subsection{Learning that a language is infinite}

One of the most striking properties of natural language is that there are infinitely many sentences of valid, grammatical English \cite<though see>{pullum2010recursion}. This fact might be considered to be a core aspect of our innate linguistic endowment, perhaps a consequence of a generative capacity that forms UG. On the other hand, once we consider learners who are computationally sophisticated, it may be possible to \emph{learn} that the generative system supporting language is infinite. After all, there are aspects of language which are \emph{not} infinite (e.g. the set of function words) and there have even been argued to be languages with finite cardinality \cite{futrellcorpus}. Interestingly, from a learning point of view, it is not entirely obvious what finite data might convince learners that their language permits infinitely many strings.

\begin{figure}
    \centering
	\includegraphics[width=0.5\textwidth]{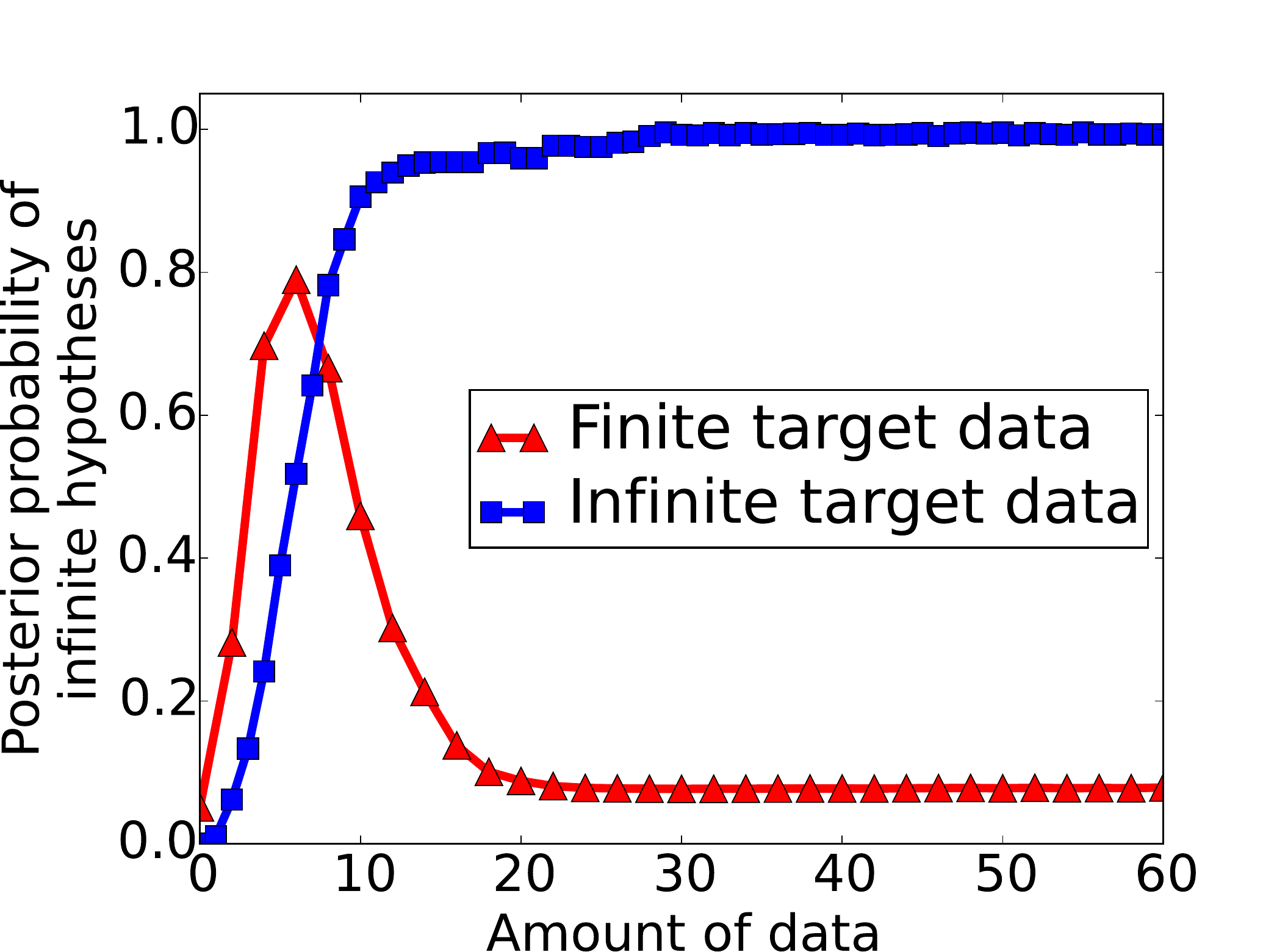}
    \caption{The sum of posterior probabilities of infinite hypotheses given finite and infinite target data.}
    \label{g_f_an}
\end{figure}

To investigate this, we generated data from two target hypothesis languages: (a) $a^n$ with $n \leq 3$ or (b) $a^n$ with $n=1,2,3,4,\ldots$. In both cases, we assume $P(a^n) \propto (\frac{1}{2})^n$. Fig.\ref{g_f_an} shows under each type of data the posterior probability of a hypothesis that generates an infinite language. The type of hypothsis the model learns in the infinite case looks like the example above for $a^n$; in the finite case, the model could learn a hypothesis like
\begin{center}
\begin{tabular}{lll}
\begin{lstlisting}[mathescape]
F1(x) = pair(a, pair(if(flip(0.5), $\emptyset$, a), if(flip(0.5), $\emptyset$, a)))
\end{lstlisting}
\end{tabular}
\end{center}
What should be clear is that as finite languages become more complex or have greater cardinality, they become more complex. In the absence of overwhelming data, learners will prefer a more concise hypothesis and this will tend to bias them towards hypotheses that generate infinite sets. It takes additional data to convince a simplicity-driven model that long strings ($n\geq 4$) are not permitted. This shows that languages with infinite cardinality may result from core infernetial properties of a computationally sophisticated learner \cite<see also>{piantadosi2017infinitely}.

\section{Learning a simplified English phrase structure grammar}

Of course, the problem faced in natural syntax is not just a series of unrelated formal languages, but data that combines many forms of computation together. We next test the learning model with a grammar of a simple subset of English:
\begin{center}
\begin{tabular}{lcl}
$S$ & $\overset{4}{\to}$ & $NP\; VP$ \\ 
$NP$ &$\overset{2}{\to}$ & $n$ \\ 
$NP$ &$\overset{1}{\to}$ & $d\; n$ \\ 
$NP$ &$\overset{1}{\to}$ & $d\; AP\; n$ \\ 
$AP$ &$\overset{3}{\to}$ & $a$ \\ 
$AP$ &$\overset{1}{\to}$ & $a\; AP$ \\ 
$VP$ &$\overset{2}{\to}$ & $v$ \\ 
$VP$ &$\overset{1}{\to}$ & $v\; NP$ \\ 
$VP$ &$\overset{1}{\to}$ & $v\; that\; S$ \\ 
\end{tabular}
\end{center}
Here, the number over each arrow gives the unnormalized probability of each expansion (so $AP$ is three times more likely to expand to ``a'' than to ``a AP''). This target grammar includes many interesting syntactic structures of English, including multiple expansions of a nonterminal type (e.g. NP), tail recursion (in AP), transitive and intransitive verbs (in VP), and sentential embedding (in VP). 

Importantly, although we have written this as a phrase structure grammar, as above, the learning model is \emph{not} told that the strings it observes come from this grammar, or indeed from any kind of phrase structure at all. The strings that are provided have no parsing information or structure: for instance, one string the model might observe is 'if d a n v then n v that d n v d n', which corresponds to the part-of-speech sequence for a sentence like ``If an angry man smiles then Jane believes that the instrument is an accordion.'' The challenge for the model is to take these kinds of part-of-speech sequences and infer the computational process out of all compositions of primitives. 

\begin{figure}[htb]
    \centering
    \hspace{-1cm}
    \begin{subfigure}[]{0.55\columnwidth}
            \includegraphics[width=\textwidth]{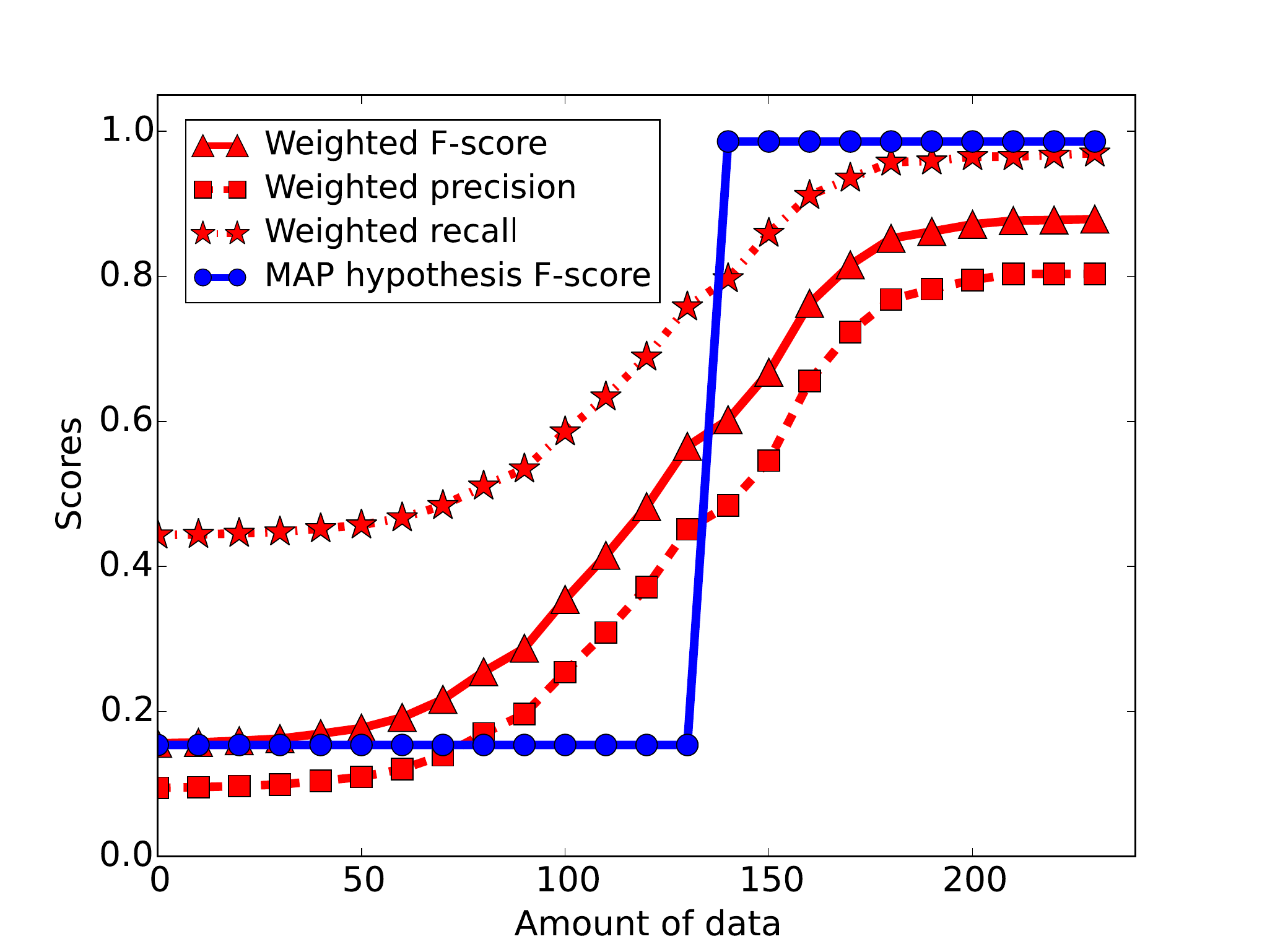}
            \caption{}
            \label{simple_curve}
    \end{subfigure}
    \hspace{-1.5cm}
    \begin{subfigure}[]{0.5\columnwidth}
            \begin{tabular}[b]{cc}
        	\begin{lstlisting}[mathescape,basicstyle=\ttfamily\fontsize{7}{6}\selectfont]
        def $F$():
            def $F_1$():
        	if $flip$():
    		    return $pair$($\mb{a}$,$recurse$())
        	else:
    	 	    return $pair$($\mb{a}$,$\mb{n}$)
        	 	
            def $F_2$($g_1$): 
                if $flip$():
                    return $\emptyset$
                else:
            	    return $pair$($\mb{d}$,$rest$($g_1$()))
            
            def $F_3$($g_1$,$g_2$): 
            	return $rest$($g_1$())
            
            def $F_4$($g_1$,$g_2$,$g_3$): 
            	$x_0$ = $pair$($pair$($\mb{d}$,$g_3$($g_1$,$g_2$)),$\mb{v}$)
            	$x_1$ = $pair$($x_0$,$\mb{t}$)
            	
                if $empty$($g_2$($g_1$)):
            	    return $pair$($x_0$,$g_2$($g_1$))
                else:
            	    return $pair$($x_1$,$recurse$($g_1$,$g_2$,$g_3$))
            	
            return $F_4$($F_1$,$F_2$,$F_3$)
        	\end{lstlisting}
        	\end{tabular}
        \vspace{.1cm}
    	\caption{}
        \label{simple_example}
    \end{subfigure}
    \caption{(a)The graph shows the weighted F-score of hypotheses learning simple English case given different sizes of data; (b) Example hypothesis for simple English.}
    \label{simple}
\end{figure}

The learning curves are shown in Figure.~\ref{simple_curve} which gives the posterior probability (y-axis) of various types of grammars as a function of amount of data (x-axis). The model eventually settles on a hypotheses Fig.\ref{simple_example}. Importantly, the components of this grammar--recurive, tail-recursive, random-choices of expansions, etc. are not ``built in'' as necessary mechanisms for learners. Instead, the parts of this grammar which mimic a PCFG are constructed by composing function calls with the assumed primitives. 

\section{Connections to human learning}

This computational model is interesting not only as a theoretical demonstration of the power of statistical learning, but also because its underlying principles can be directly connected to human behavior. In this section, we consider two prior experiments on human artificial grammar learning (AGL) and show that the model exhibits similar patterns as humans. 

\subsection{Structure of the input}

\begin{figure}[tb]
	\centering
    
	\begin{subfigure}[]{0.3\textwidth}
    \centering
	\includegraphics[width=\textwidth]{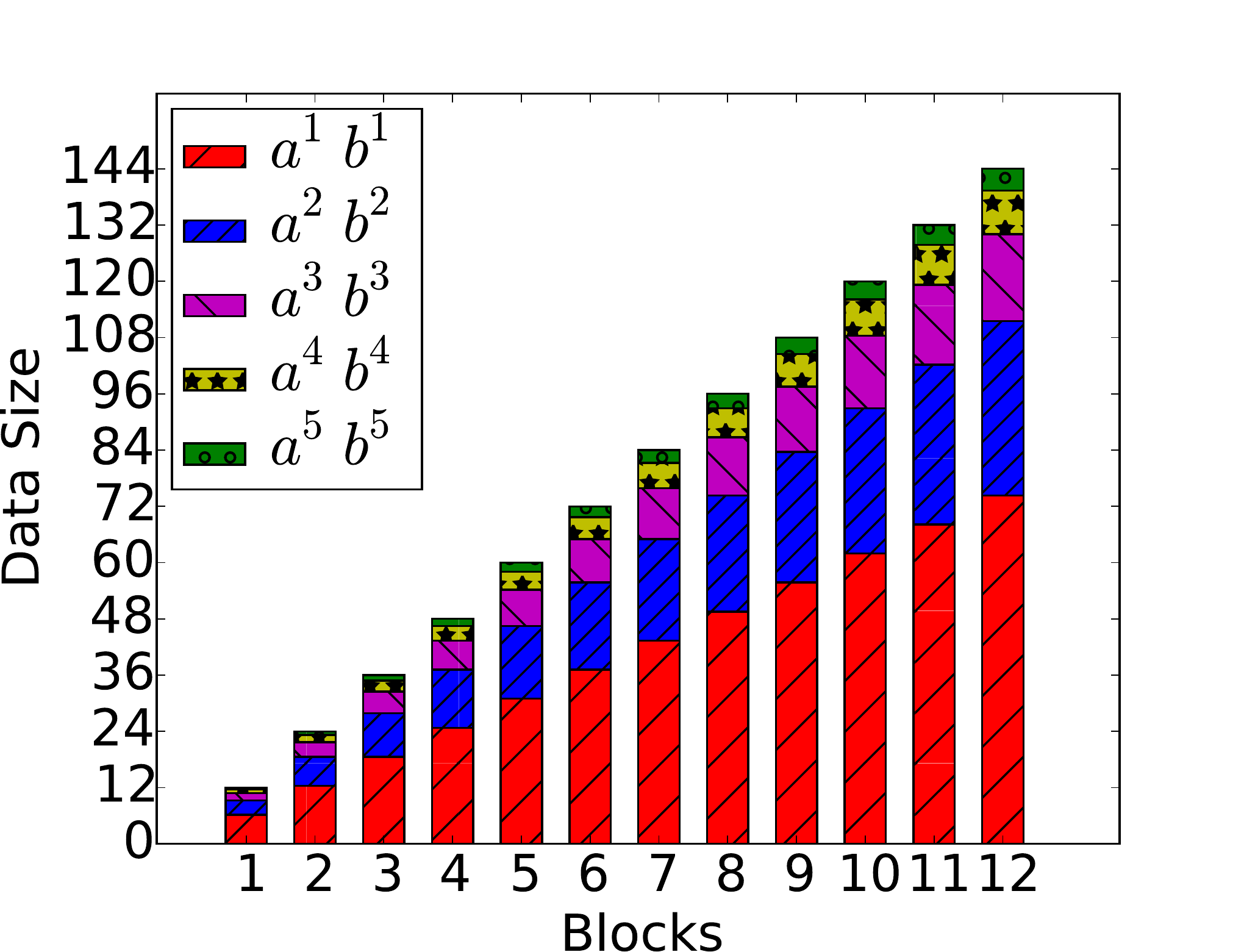}
    \caption{}
    \label{skewed_g}
	\end{subfigure} \hfill
    \begin{subfigure}[]{0.3\textwidth}
    \centering
	\includegraphics[width=\textwidth]{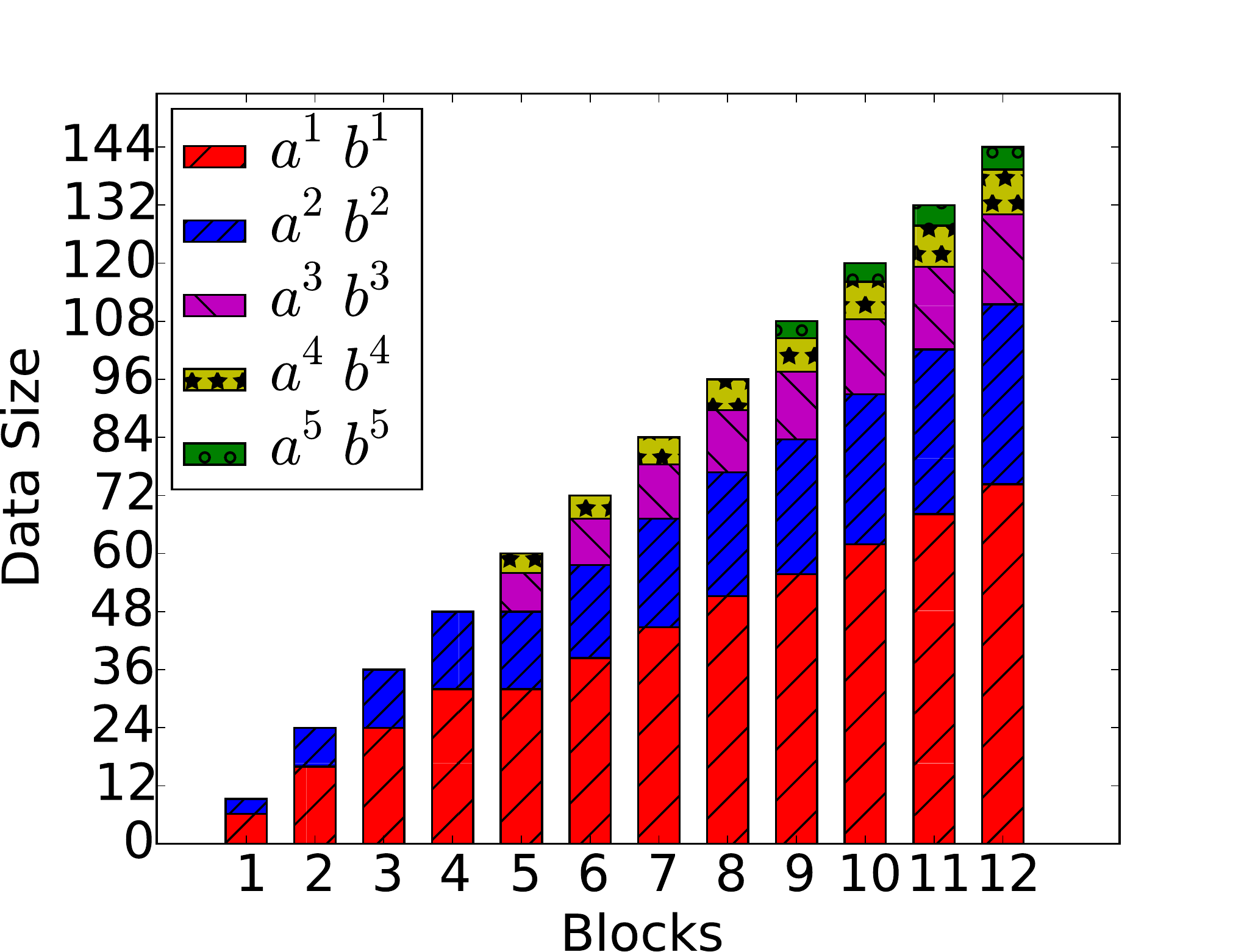}
    \caption{}
    \label{staged_g}
	\end{subfigure} \hfill
     \begin{subfigure}[]{0.3\textwidth}
    \centering
	\includegraphics[width=\textwidth]{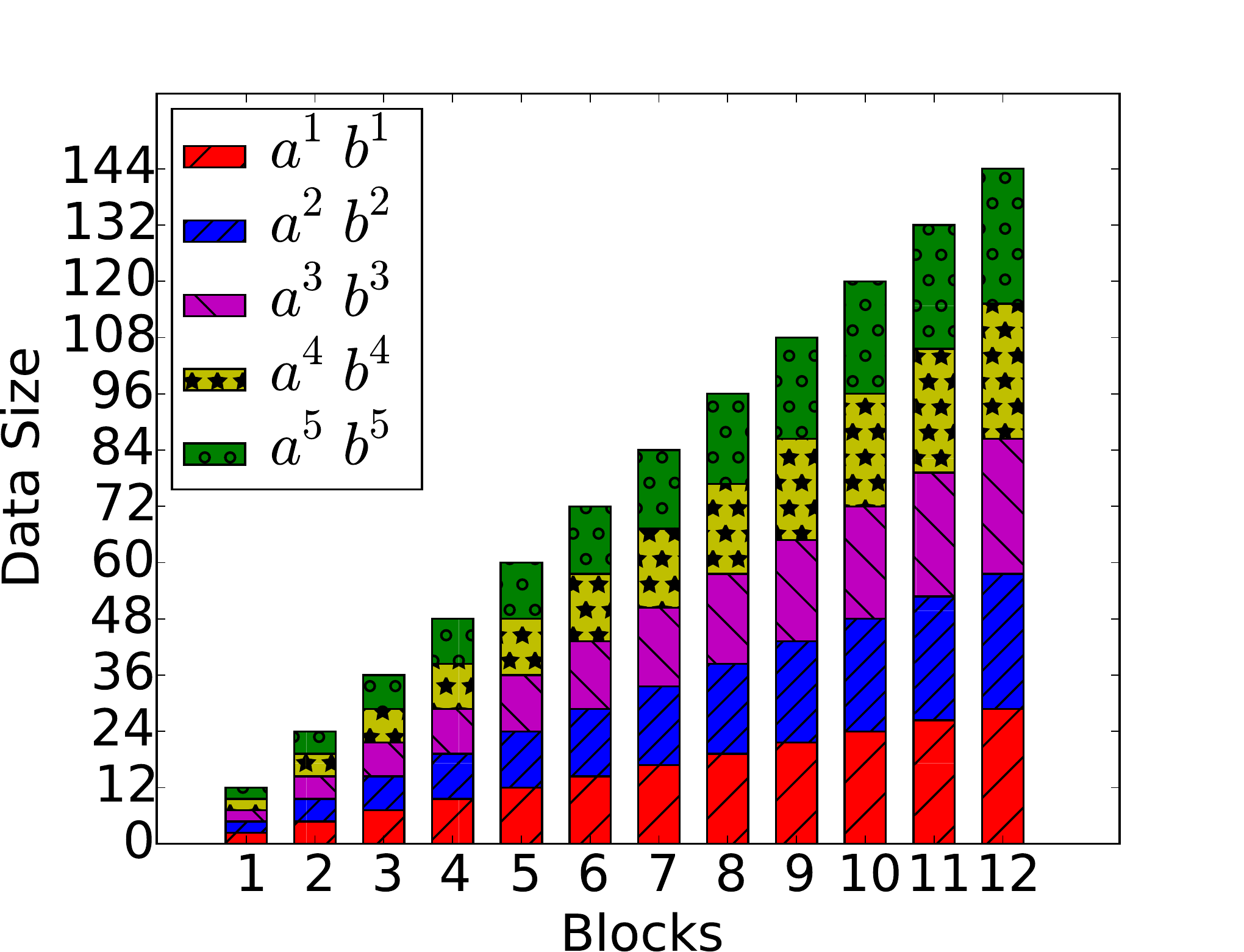}
    \caption{}
    \label{random_g}
    \end{subfigure}
    
    \caption{The compositions of data blocks in experimental conditions from Lai et al (2011, 2013): (a) skewed frequency, (b) staged input, and (c) random.}
    \label{fig:data_dist}
\end{figure}

A wealth research has sought to understand human of learning center-embedded recursion structure~\cite{fitch2004computational, lai1, lai2}. This structure, in our context, can be characterized as an ``$a^nb^n$'' language, and an example in natural language can be ``the car that I drove yesterday broke''. It is of particular interest because it is shown to be sufficiently complex and is a crucial feature in human language faculty~\cite{hauser}. \citeauthor{lai2} studied whether an unequal distribution of the training exemplars can influences the process of learning a center-embedded grammar. They argued that a biased distribution, where short exemplars are more likely to be presented to the learners than longer ones, can facilitate acquisition. In Lai et al.'s experiments, participants were asked to learn an artificial language generated with center-embedded recursive grammar which resembles the $a^nb^n$ language. They were exposed to 12 blocks of training data set and completed a test at the end of each block. Participants were divided into two groups. Training data for the first group was randomly sampled from the grammar (Random Case), where short and long exemplars has the same amount; the one for the other group has a biased distribution over exemplars. The amount of exemplars rises exponentially as they get shorter, e.g., the shortest exemplars are 2 times more than the second shortest and 4 times than the third shortest, and so on. These settings are illustrated in Fig.\ref{skewed_g} and Fig.\ref{random_g}. Their results showed an improvement in learning as measured by the rate of correctly recognizing the target language in test phrase: when given the data with biased distribution, participants performance increased by 30\%.

\citeA{lai1} studied another factor that can facilitate the learning, i.e. staged input. They argued that AGL can be facilitated with an explicit control of exemplar difficulties in several learning stages. The experiments setup was similar to the previous one, except that learners in the test group now experience 3 stages of training, each containing 4 blocks. For each stage, the maximum length (or complexity) of exemplars was limited, and this limit increased as participants proceed to higher stage. This setting is illustrated in Fig.\ref{staged_g}. Experimental results showed improvement in performance by 40\% when learners were exposed to staged training.

We constructed three datasets of the same size (144 items) and language ($a^n b^n$) as that used in these experiments. We matched the input data composition to these studies. The only difference was the actual appearance of exemplars to be presented, which for simplicity we mapped to ``a'' and ``b'' instead of syllables. 

\begin{figure}[htb]
	\centering    
	\includegraphics[width=.4\textwidth]{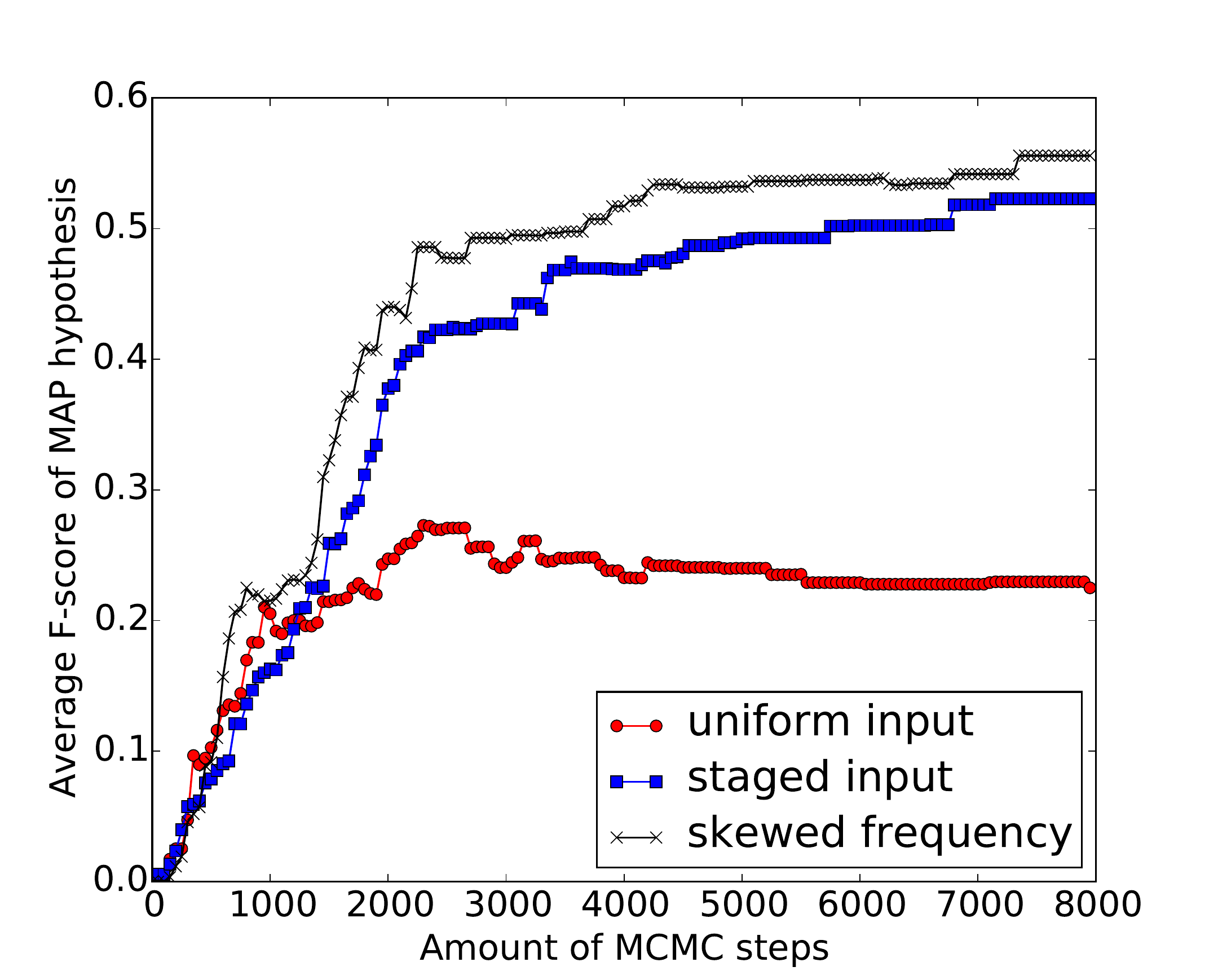}
    \caption{The average F-score of MAP hypothesis given different types of data (skewed frequency, staged input and uniform input) as a function of the total MCMC steps taken.}
    \label{fig:si_sf_un}
\end{figure}

Figure Fig.\ref{fig:si_sf_un} shows the learning curves for each of these kinds of input. These results show that models with skewed frequency and staged input performs better than the random case---reaching a higher F-score faster---agreeing qualitatively with the human results. The computational explanation for this is also intuitive: data with a uniform distribution requires an equal chance to produce strings with every different level of complexity, which is hard to realize in our model: the main device for producing complex strings that follows a certain type of grammar is stochastic recursion, and that the chance for generating complex strings decreases exponentially is almost an innate property of this mechanism.

\subsection{Nonadjacent dependency}

The third case we consider is that of nonadjacent dependencies, one of the most characteristic features of human languages. Examples can be found in the dependencies between auxiliaries and inflectional morphemes (e.g. ``\ul{is} runn\ul{ing}, \ul{has} eat\ul{en}''), as well as dependencies involving number agreement (``The rock\ul{s} on the bluff \ul{are} jagged''); nested nonadjacent dependencies can be found in, for instance, the if-then constructions (above) \cite{chomsky1957syntactic}. 

In the language learning literature, dependencies have been studied in very simple constructions without recursive embedding. \citeA{gomez} had participants learn an $aXb$ like language, meaning that the first ($a$) and last ($b$) word were fixed and $X$ was free to vary. They found that it was \emph{easier} to for people to discover the fixed dependency between $a$ and $b$ if the $X$ is \emph{more variable}. In their experiment, $X$ was chosen from a fixed pool of strings; as the cardinality of the pool increased, learners performed better. The result is intuitive: when there are more possible $X$es it is less likely to be predicted since there are more possibilities may free resources for discovering the association between $a$ and $b$.

\begin{figure}[htb]
    \centering
    
    \begin{subfigure}[]{0.4\textwidth}
    \centering
	\includegraphics[width=\textwidth]{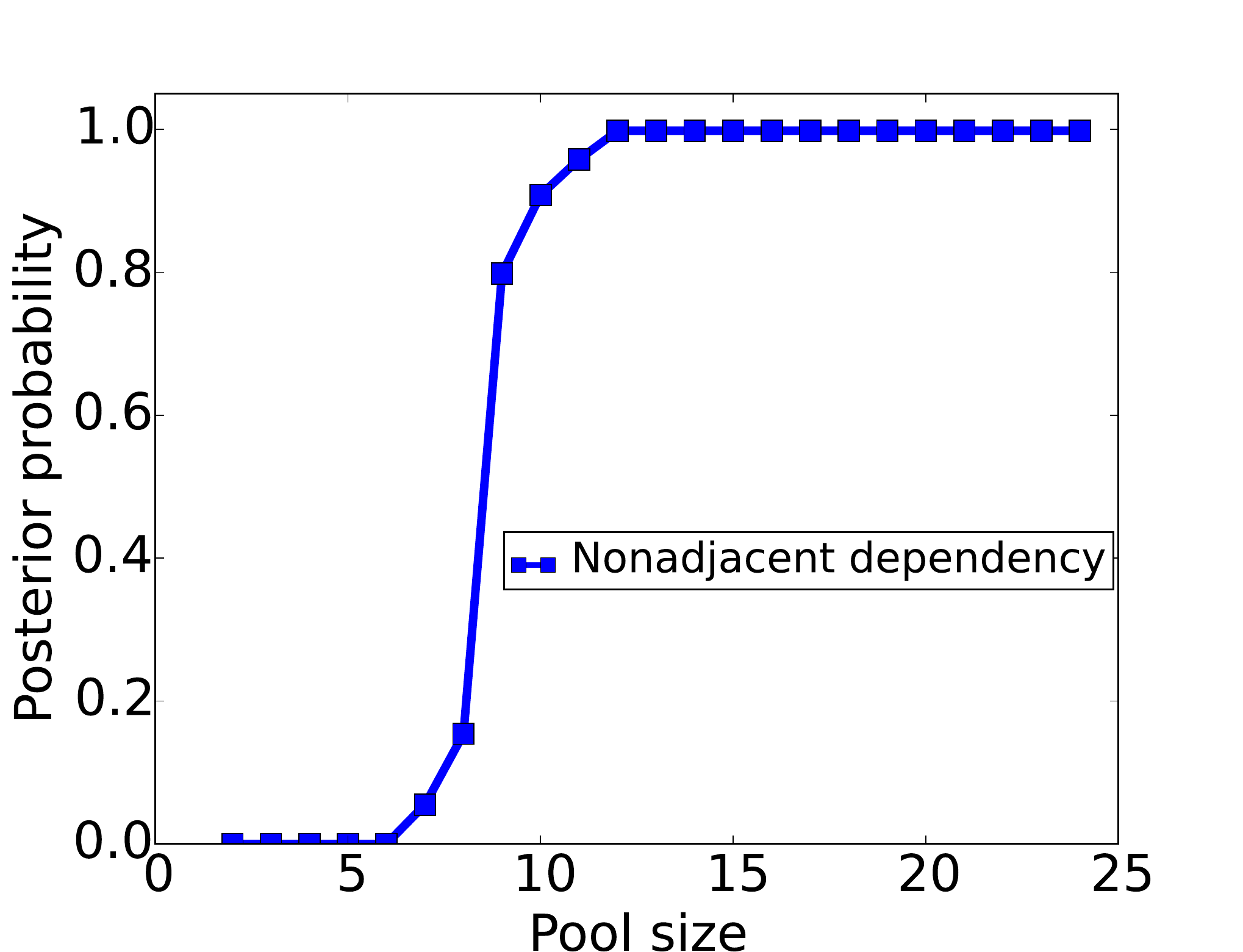}
    \caption{}
    \label{non_g}
	\end{subfigure}
    \hspace{-1.6cm}
	\begin{subfigure}[]{0.3\textwidth}
    \centering
	\begin{tabular}[b]{cc}
        \begin{lstlisting}[mathescape,basicstyle=\ttfamily\fontsize{7}{6}\selectfont]
        Small pool size
        
        def $F$():
            def $F_1$():
                if $flip$():
                    return $pair$($\mb{a}$,$pair$($\mb{h}$,$pair$($\mb{h}$,$pair$($\mb{h}$,$\mb{b}$))))
                else $flip$():
                    return return $pair$($\mb{a}$,$pair$($\mb{h}$,$pair$($\mb{h}$,$pair$($\mb{i}$,$\mb{b}$))))
                    
            def $F_2$($g_1$):
                return $g_1$()
            
            return $F_2$($F_1$)
        \end{lstlisting}
        \end{tabular}
    \caption{}
    \label{non_example}
	\end{subfigure}
    \begin{subfigure}[]{0.3\textwidth}
    	\centering
        \vspace{-1.4cm}
		\begin{tabular}[b]{cc}
        \begin{lstlisting}[mathescape,basicstyle=\ttfamily\fontsize{7}{6}\selectfont]
        Large pool size
        
        def $F$():
            def $F_1$():
                x = $\mb{h}$
                for _ in xrange(3):
                    x = $pair$(x,$\mb{h}$) if $flip$() else $pair$(x,$\mb{i}$)
                return x
                
            def $F_2$($g_1$):
                return $pair$($\mb{a}$,$pair$($g_1$(),$\mb{b}$))
            
            return $F_2$($F_1$)
        \end{lstlisting}
        \end{tabular}
	\end{subfigure}
    \caption{(a) The graph shows the weighted F-scores of adjacent and nonadjacent dependencies, and their differences with different pool sizes; (b) example hypotheses with small and large pool sizes (only one pair of dependency is shown for simplicity).}
    \label{non}
\end{figure}

We adopt the same paradigms in computational test. There are 3 pairs of nonadjacent dependencies. They are $aXb$, $cXb$ and $cXd$, where $X$ is a string with its length equals 3, and the characters in $X$ are $h$, $i$, $j$, $k$. We construct the pool that $X$ is drawn from by numerating all possible 3-characters strings' permutations (the total amount is, therefore, $3^4=81$). When generating the strings, we draw $X$ uniformly from the pool.

Gomez et al. obtained their results by showing learners strings with both trained and untrained dependencies (ones that do not appear in training strings), and then ask them to choose whether string is valid or not. Then the rates of learners accepting trained and untrained exemplars are computed respectively, and the difference of these two values is taken as the final measure of performance. In our computational test, we can keep track of how much attention our model puts to the nonadjacent dependencies by looking at the posterior probability of those hypotheses that generate such structures. The idea that links these two measures are straightforward: if learner assigns more posterior weight on correct hypotheses, then the probability of generating the untrained data given the posterior distribution state would be smaller, thus less chance to accept them as valid strings.

To get this measure, we first evaluate every hypothesis in our finite hypothesis set for 2048 runs, and check if all generated strings contain valid pairings, such as $aXb$ and $cXd$. In this case, those candidates with all their strings valid are taken as the hypotheses that contain the nonadjacent dependencies. Then we compute the posterior of those hypotheses given different pool size ranging from 2 to 25. The posterior probability curve is shown in Fig.\ref{non}.

This result reveals a similar learning preference as we observed in finite vs. infinite language case. When target language has a small cardinality (i.e. small pool size), the model tends to memorize all occurred strings (examples shown in Fig.\ref{non_example}). Since by this time, the cost of memorizing it still remains low---lower than that of exploring the underlying generative model of the target language. But as the pool size grows, memorizing observed data starts to increase the model complexity. So model tends to explore the potential rules for generating the data, leading to a increase in the posterior probability of those hypotheses containing nonadjacent dependencies.

\section{General discussion \& Conclusion}

Statistical language learning has often been criticized by formal linguists as failing to do justice to the structure of language. Indeed, the earliest statistical learning paradigms \cite{saffran1996statistical, aslin1998computation, saffran1999statistical} focused on domains with little structure such as word segmentation. But this was for good reason: the computational system behind syntax and structure is complex, and a considerable amount is learned about the general capacity of human learners---what general class of theories should science explore---by demonstrating their statistical acquisition in any domain. Indeed, these early statistical learning experiments motivated the development of richer computational models \cite<e.g.>{redington1998distributional}, as well as \citeA{perfors2011learnability}'s work that captured learners as doing model comparison. Perhaps the most interesting critique of Perfors et al. has argued that it ``builds in'' \emph{more} than alternatives such as Chomsky's \emph{Universal Grammar} (UG) because in Perfors et al.'s work, learners consider seven different grammars. Surely it is a more parsimonious theory of human nature that learners come equipped with one single grammar than seven. The arguement also applies to Chater et al.: they allow learners to consider an infinite number of grammars. How could ``building in'' an infinite number of grammars be a more plausible or parsimonious theory than building in just one?

It turns out that this thinking is wrongheaded---but for a very interesting reason. The argument can best be understood by considering Jorge Luis Borges' short story, \emph{The Library of Babel}. Borges imagines an infinite library full of every possible book---every possible sequence of characters. The curious fact about the library which contains \emph{every} possible book is that it actually contains \emph{essentially no information at all.} Its contents can be completely specified with just a few words (``all possible books''), or compressed into an extremely short generating computer program, yielding a negligible complexity or description length. Certainly, it is much easier to describe the entire library of Babel than to describe one of its typical books. In this way, it is often more concise or parsimonious to build in a larger, unconstrained space of hypotheses because a large space can easily have a more concise description (e.g. ``all computations'') compared to a constrained hypothesis space. 

The model described in this paper shows in principle how learners can build in very little in terms of representation---just the specific operations in Table 1. This model is the first to show that statistical learning can \emph{construct} representations of provably different computational power, including regular, context-free, and context-sensitive hypotheses and in doing so, it requires surprisingly little data\footnote{There is an interesting sense in which models such as these challenge the very notion of innateness. We could ask: are the infinitely many hypotheses considered here ``innate''? Or is only the \emph{capacity for generating} hypotheses innate? We tend toward the latter sense because the amount that is ``built in'' is very small. When we download a C compiler, it would be very strange to think that all of the programs that could be written are ``innately'' included. Instead, the compiler has a generative capacity to evaluate many such programs, and it is left up to the programmer to construct the specific program that should be run.}.




The overarching view, then, is not one of comparing a few different built-in grammars, but applying likely the same remarkable capacity for discovering algorithms that is apparent in other cognitive domains. As the recent innovations in the Bayesian LOT and inductive computation \cite{chater2007ideal, hsu2010logical, hsu2011probabilistic} make clear, such statistical models are just getting started. However, is likely that ongoing research will push towards more detailed linguistic facts and correspondingly sophisticated learning theories.\footnote{Recently, for example, \citeA{mitchener2010computational} studied \citeA{berwick2011poverty}'s example ``*The child seems sleeping'' and showed that a hierarchical Bayesian model motivated by Perfors et al. and other models showed promise in acquiring the verb's syntax.} The work on the simplified English grammar has shown that the same model can begin to scale to more complex types of computations and syntactic processes.

\section{Conclusion}

We have shown that a statistical learner who operates over computational processes can discover many of the key features of syntactic structures, including regular, context-free, and context-sensitive langauges, using positive evidence only and a simple, unitary learning mechanism. This work opens the door to more sophisticated statistical learning theories that incorporate other areas of language like semantics and pragmatics into the generating process, and unifies theories of syntactic learning with state of the art machine learning models used in other areas of cognition. 

\bibliographystyle{apacite}
\bibliography{ref}

\end{document}